\documentclass[conference]{IEEEtran}
\IEEEoverridecommandlockouts

\usepackage{hyperref}
\usepackage{float}
\usepackage{ifthen}
\usepackage{tikz}
\usepackage{makecell}
\usetikzlibrary{positioning}
\usepackage{cite}
\usepackage{amsmath, amssymb, amsfonts}
\usepackage{graphicx}
\graphicspath{ {./images/} }
\usepackage{textcomp}
\usepackage{xcolor}
\usepackage{tabularx, booktabs}
\usepackage{textgreek}
\usepackage[ruled, longend]{algorithm2e}
\usepackage{tcolorbox}
\usepackage{subcaption}

\DeclareMathSymbol{\shortminus}{\mathbin}{AMSa}{"39}
\newcolumntype{C}{>{\centering\arraybackslash}X}

\def\BibTeX{{\rm B\kern-.05em{\sc i\kern-.025em b}\kern-.08em
    T\kern-.1667em\lower.7ex\hbox{E}\kern-.125emX}}

\begin{document}

\title{Spatio-Temporal Look-Ahead Trajectory Prediction using Memory Neural Network}

\author{ 
        \IEEEauthorblockN{Nishanth Rao}
        \IEEEauthorblockA{\textit{Dept. of Aerospace Engineering} \\ \textit{Indian Institute of Science} \\ Bangalore, India \\ nishanthrao@iisc.ac.in} \and
        
        \IEEEauthorblockN{Suresh Sundaram}
        \IEEEauthorblockA{\textit{Dept. of Aerospace Engineering} \\ \textit{Indian Institute of Science} \\ Bangalore, India \\ vssuresh@iisc.ac.in}
        }

\maketitle

\tikzset{%
	every neuron/.style={
		circle, 
		draw,
		minimum size = 0.7cm
	},
	neuron memory/.style={
		circle,
		draw,
		minimum size = 0.5cm, 
		fill=black
	},
	neuron missing/.style={
		draw=none,
		scale=2.5,
		text height=0.133cm,
		execute at begin node=\color{black}$\vdots$
	},
}

\begin{abstract}
Prognostication of vehicle trajectories in unknown environments is intrinsically a challenging and difficult problem to solve. The behavior of such vehicles is highly influenced by surrounding traffic, road conditions, and rogue participants present in the environment. Moreover, the presence of pedestrians, traffic lights, stop signs, etc., makes it much harder to infer the behavior of various traffic agents. This paper attempts to solve the problem of spatio$\shortminus$temporal look$\shortminus$ahead trajectory prediction using a novel recurrent neural network called the Memory Neuron Network. The Memory Neuron Network (MNN) attempts to capture the input-output relationship between the past positions and the future positions of the traffic agents. The proposed model is computationally less intensive and has a simple architecture as compared to other deep learning models that utilize LSTMs and GRUs. It is then evaluated on the publicly available NGSIM dataset and its performance is compared with several state$\shortminus$of$\shortminus$art algorithms. Additionally, the performance is also evaluated on a custom synthetic dataset generated from the \texttt{CARLA} simulator. It is seen that the proposed model outperforms the existing state$\shortminus$of$\shortminus$art algorithms. Finally, the model is integrated with the \texttt{CARLA} simulator to test its robustness in real$\shortminus$time traffic scenarios.
\end{abstract}

\section{Introduction}
Research in autonomous vehicles has attracted a lot of interest from researchers around the world. With the rise of electric vehicles over the past few years, autonomous navigation and path planning have become an inherent feature of these vehicles. In presence of traffic, these vehicles should reach their destination and also follow traffic rules, prevent accidents, detect various traffic signs, handle reckless drivers and rogue vehicles. To be able to perform the aforementioned tasks, the autonomous vehicle must have the ability to predict the motion of it's surrounding vehicles. This will enable the vehicle to make necessary decisions at the right time. Anticipating traffic scenarios is thus a major functionality of autonomous vehicles in order to navigate safely amidst their human counterparts.

This is a very challenging problem due to the unpredictable nature of traffic agents. Their behaviour is often determined by multiple latent variables that cannot be estimated beforehand in new and unknown environments, such as the mental state and driving experiences of human drivers, road and weather conditions, destination of each vehicle in the traffic, reckless behaviour of traffic agents that involve overtaking, abrupt lane changing without indication, etc. 

Many recent state$\shortminus$of$\shortminus$art deep learning models have utilized \textit{Long-Short Term Memory} (LSTM) networks\cite{hochreiter1997long} and \textit{Gated Recurrent Units} (GRUs)\cite{cho2014learning} for the trajectory prediction problem. One technique that is utilized by many approaches is that of an encoder-decoder architecture. In these approaches, the spatio-temporal context from the vehicle trajectories is extracted and then a recurrent neural network (RNN) based decoder is used to predict the future trajectories. While they have been successful in regressing the future trajectories of traffic agents over a certain time horizon, they are heavily dependent on computational resources due to their complex architecture and require a lot of training time.

This paper will attempt to address all the aforementioned problems by adapting a unique recurrent neural network called the Memory Neuron Network\cite{sastry1994memory}. The Memory Neuron Network is an extension of the traditional neural network with addition of memory elements to each neuron in the network, that are capable of storing temporal information. This network has a simple architecture, and requires less computational resources as compared to the currently available state$\shortminus$of$\shortminus$art deep learning methods. The performance of the proposed model is evaluated on the publicly available NGSIM US-101 dataset. Although, NGSIM dataset provide comprehensive data of real traffic agents, it does not contain sufficient data for reckless and rogue traffic agents. To address this situation, a synthetic dataset is generated using the \texttt{CARLA} simulator\cite{Dosovitskiy17} that contains the trajectories of multiple heterogeneous rogue traffic agents. As the proposed model is computationally less intensive, it allows for the deployment onto all the rogue vehicles present in the real$\shortminus$time traffic simulation with additional 80 normal cars. To summarize, our main contributions are as follows:
\begin{itemize}
    \item A novel model is proposed that uses a recurrent neural network $\shortminus$ the Memory Neuron Network for the problem of spatio$\shortminus$temporal look$\shortminus$ahead trajectory prediction.
    \item The proposed model is evaluated on publicly available US$\shortminus$101 dataset, and the RMSE is reported along with several state$\shortminus$of$\shortminus$art methods.
    \item To evaluate the performance of our model with respect to reckless drivers, rogue vehicles are simulated on \texttt{CARLA} simulator and their trajectories are recorded. The model is then implemented in real$\shortminus$time simulation on each rogue vehicle with a look$\shortminus$ahead horizon of $5s$, demonstrating the robustness and the computational efficiency of the proposed model.
\end{itemize}

\section{Related Work}
This section sets out to explore some of the various methods currently present in the literature to address the motion prediction problem. The existing literature can be broadly classified into three parts which are discussed below.
\subsection{Mechanics-based methods}
In these approaches, vehicles are mathematically modelled using Newtonian laws of translation and rotation. Once the model is formed, an Unscented Kalman Filter (UKF) is used to estimate the states of the vehicles. \cite{xie2017vehicle} propose an Interactive Multiple Model Trajectory Prediction (IMMTP) which combines physics-based and manoeuvre-based predictive models. \cite{veeraraghavan2006deterministic} use a deterministic sampling approach in the UKF process for a robust estimate of target trajectories. These models work really well in certain scenarios and short time prediction horizon. However, these approaches tend to linearize the obtained models and hence, are unable to capture the inherent non-linear characteristics in a generic traffic scenario. Another issue with these approaches is that the parameters of the mathematical model such as the dimensions of the vehicle, its braking coefficients, steering torque etc., must be set and tuned in real$\shortminus$time, as soon as a vehicle is detected in the vicinity. This may not be feasible when the other agent's model is unknown.  A detailed study on these methods can be found in \cite{schubert2008comparison}.
\subsection{Human behavior-based models}
These techniques attempt to build a mathematical formulation of the human behavior and utilize these as a model for the driving process. \cite{li2016human} apply the \textit{theory of planned behavior} to model the driver behavior, and develop a driver model that accounts for various human aspects such as driving experiences, emotions, age, gender etc. \cite{ferreira2013gender} and \cite{puterman2014markov} apply control theory and Markov Decision Process (MDP) to model human behaviors specifically for the navigation process in a single lane. To extend the analysis to multi$\shortminus$lane junctions, Hidden Markov Models are proposed to model human behaviors in \cite{zou2006modeling}. Statistical models have been proposed in \cite{boyraz2009driver}, \cite{dapzol2005driver} and \cite{ziebart2009human} to predict driving manoeuvres and behaviors. These methods work best when the knowledge of the human behaviors and their analysis are known beforehand. However, in the case of new and unknown environments these models fail to provide reliable predictions.

\subsection{Deep learning methods}
These methods use a spatial encoder to process the raw trajectory data, and then use recurrent neural networks to estimate the future trajectories. To extract the spatial context from the trajectories, \cite{gupta2018social}, \cite{vemula2018social} and \cite{qi2017pointnet} use a sequential point$\shortminus$based representation. Occupancy grid$\shortminus$base is another popular representation for the spatial context. These approaches model trajectories as a $2D$ sequence, which can be unstructured at times due to the missing temporal information. Extraction of the temporal context is normally done by using RNNs. \cite{subhrajit2020bayesian} propose a Bayesian fuzzy model to accurately estimate the temporal dependencies. \cite{li2019grip} and \cite{nikhil2018convolutional} also use Convolutional Neural Networks to encode the temporal context. To unify the spatial and temporal contexts, \cite{he2020ust} follows a simple and effective approach, where both the contexts are encoded together, using a Multi-Layer Perceptron, which drastically improves the prediction performance. \cite{messaoud2020attention} use a RNN based encoder-decoder along with \cite{bahdanau2014neural} to model the spatio-temporal context. For the process of predicting future trajectories different variants of RNNs have been used. \cite{altche2017lstm} use a standard LSTM network for trajectory prediction on highways. \cite{bhattacharyya2018multi} and \cite{si2019agen} use Imitation Learning along with Generative Adversarial Networks to predict future trajectories. \cite{deo2018convolutional} use LSTMs along with Convolutional Neural Networks with social pooling layers and generate a multi-modal Gaussian model for trajectory prediction. While these approaches have helped in improving the performance, they require heavy computational resources. This can make them quite hard to be implemented in real$\shortminus$time scenarios. 

\section{Trajectory Prediction Framework}
Fig. \ref{MNN} shows the proposed model for trajectory prediction. The figure consists of a \textit{trajectory database}, that consists of all the change in trajectory samples for multiple vehicles, present in the dataset, and the Memory Neuron Network which is shown as a black box. At every time instant $t$, the trajectory database provides the change in the $(x, y)$ coordinates for a particular vehicle, and the network estimates the next change in position of the vehicle. The initial values provided by the trajectory database is fed to the network multiple times sequentially, so that the predicted values reach a steady$\shortminus$state. Once the steady$\shortminus$state is achieved, the network then receives consecutive input values from the trajectory database.
\begin{figure}
\begin{tikzpicture}[x=1.5cm, y=1.5cm, >=stealth]

\draw (0.6, 3.5) rectangle (3.1, 5.0) node[pos=.5]{\makecell[l]{Trajectory \\ \ Database: \\ $\Delta x_1, \Delta x_2, ...$ \\ $\Delta y_1, \Delta y_2, ...$}};

\draw (3.1, 4.6) -- (3.7, 4.6) node[above, midway]{$\Delta x_{t}$};
\draw (3.1, 4.0) -- (3.7, 4.0) node[above, midway]{$\Delta y_{t}$};
\draw [->](3.7, 4.6) -- (4.2, 3.72);
\draw [->](3.7, 4.0) -- (4.07, 3.6);

\draw [fill= black!10] (1.2, 2.2) rectangle (2.9, 3.2) node[pos=.5]{\makecell[l]{Memory \\ Neuron \\ Network}};
\draw (2.9, 2.9) -- (3.7, 2.9) node[above, midway]{$\Delta \hat{x}_{t}$};
\draw (2.9, 2.4) -- (3.7, 2.4) node[above, midway]{$\Delta \hat{y}_{t}$};

\draw [->](3.7, 2.9) -- (4.07, 3.4);
\draw [->](3.7, 2.4) -- (4.2, 3.27);
\draw [->](4.5, 3.5) -- (5.1, 3.5) node[above, midway]{$e_t$};

\draw (3.0, 2.9) -- (3.0, 2.0);
\draw (3.2, 2.4) -- (3.2, 1.8);

\draw (3.0, 2.0) -- (2.4, 2.0);
\draw (3.2, 1.8) -- (2.4, 1.8);

\draw (1.9, 1.7) rectangle (2.4, 2.1) node[pos=.5]{\makecell[l]{$z^{\shortminus 1}$}};

\draw (1.9, 2.0) -- (1.0, 2.0);
\draw (1.9, 1.8) -- (0.8, 1.8);

\draw [->](1.0, 2.0) -- (1.0, 2.9) -- (1.2, 2.9) node at (0.8, 3.1) {$\Delta \hat{x}_{t\shortminus 1}$};
\draw [->](0.8, 1.8) -- (0.8, 2.4) -- (1.2, 2.4) node at (0.65, 2.55) {$\Delta \hat{y}_{t\shortminus 1}$};

\node at (3.0, 2.9)[circle, fill, inner sep=0.9pt]{};
\node at (3.2, 2.4)[circle, fill, inner sep=0.9pt]{};

\draw [fill=gray!50](4.3, 3.5) circle [radius=0.25] node{\makecell[l]{$\sum$}};
%\draw (1.7,2.9) rectangle (2.2, 3.3) node[midway]{$z^{\shortminus 1}$};

\end{tikzpicture}
\centering
\caption{Spatio-temporal lookahead model} \label{MNN}
\end{figure}
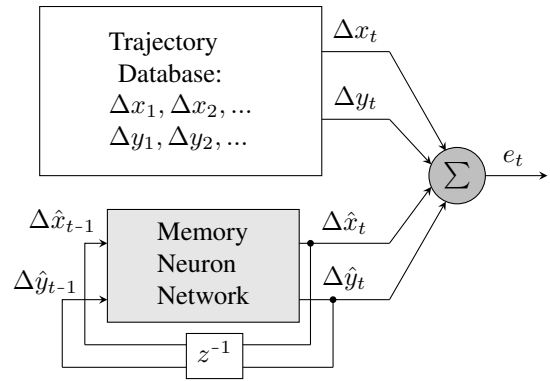
\begin{figure}
\begin{tikzpicture}[x=1.5cm, y=1.5cm, >=stealth]

\draw (0.2,0) -- (4.2, 0);
\draw (0.2, 0) -- (0.2, -3);
\draw (4.2, 0) -- (4.2, -3);
\draw [dashed] (0.2, -1.1) -- (4.2, -1.1);
\draw [dashed] (0.2, -1.9) -- (4.2, -1.9);
\draw [dashed] (0.2, -0.2) -- (4.2, -0.2);
\draw [fill=black!50] (1.5, -1.75) rectangle (2.5, -1.25); %Ego vehicle
\draw [fill=black!10] (0.2, -1.75) rectangle (0.9, -1.25); %behined EV
\draw [fill=black!10] (3.5, -1.75) rectangle (4.2, -1.25); %front EV
\draw [fill=black!10] (0.9, -0.9) rectangle (1.9, -0.4); %Top left of EV
\draw [fill=black!10] (2.7, -0.9) rectangle (3.7, -0.4); %Top right of EV
\draw [fill=black!10] (2.5, -2.6) rectangle (3.5, -2.1); %Bottom right of EV
\draw [fill=black!10] (0.5, -2.6) rectangle (1.5, -2.1); %Bottom left of EV
\draw [dashed] (0.2, -2.8) -- (4.2, -2.8);
\draw (0.2, -3) -- (4.2, -3);
\draw [->] (2.0,-1.5) -- (3.0,-1.5) node at (3.0, -1.35) {$y_t$};
\draw [->] (2.0, -1.5) -- (2.0, -2.1) node at (2.15, -2.1) {$x_t$};
\node at (2.0, -1.5)[circle, fill, inner sep=1.0pt]{};
\end{tikzpicture}
\centering
\caption{The coordinate system is shown for a particular ego vehicle in a multi-lane traffic environment. The y-axis is along the longitudinal direction and the x-axis is perpendicular to it.} \label{CS}
\end{figure}
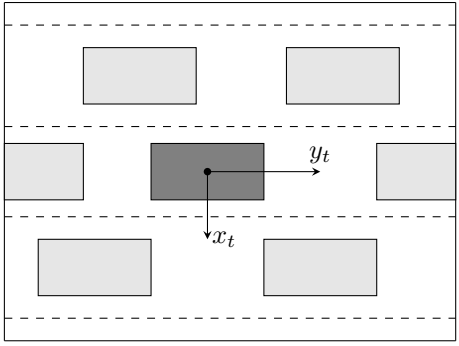
\subsection{Problem Formulation}
The coordinate system used for formulating the trajectory prediction problem is shown in Fig. \ref{CS}. It shows the ego vehicle (filled rectangle) and the non$\shortminus$ego vehicles surrounding it (hollow rectangles). The location of the vehicle is measured at its centre of mass in the local coordinate frame instead of the global coordinate frame (GPS data). The ego vehicle is assumed to be equipped with sensors that can measure the position and velocity of the surrounding non$\shortminus$ego vehicles in the local coordinate frame . In this manner, it is possible to obtain the track histories of the non$\shortminus$ego vehicles present in the vicinity of the ego vehicle.

The inherent uncertainties of the sensors only provide an approximate estimate of the position and velocities of the surrounding vehicles. As a result, it is challenging to predict the future trajectories of these vehicles using simple kinematic equations. Thus, as followed in \cite{kumar2006identification}, a \textit{data - driven} model is developed that can relate the past track histories of the vehicles to their future trajectories.
As the values of the trajectory data can change drastically when driving from one point to another over long periods of time, the difference between consecutive $(x, y)$ coordinates are taken:
\begin{equation}
	\Delta \mathbf{x}_t = \mathbf{x}_t \shortminus \mathbf{x}_{t\shortminus 1}
	\label{state_diff}
\end{equation}
where $\mathbf{x}_t = (x_{_t}, y_{_t})$ are the local coordinates of a vehicle at time instant $t$. As the datasets are generated through sampling data points uniformly, the difference in the trajectory samples will be bounded within certain limit, ensuring network stability and improved performance. The trajectory prediction problem is then, posed as a \textit{system identification} problem, with the state of the system given by $\Delta \mathbf{x}_t$. Assuming this system is \textit{observable}, from \cite{leontaritis1985input} the state of the system can be formulated as:
\begin{equation}
	\Delta \mathbf{x}_t = F(\Delta\mathbf{x}_{t\shortminus 1}, \Delta\mathbf{x}_{t\shortminus 2}, ..)
	\label{formulation}
\end{equation}
where $F(.)$ is an unknown nonlinear function of the previous states. The goal of the network is to predict the next change in coordinates ($\Delta \hat{\mathbf{x}}_t$) of the vehicle at time $t$ such that the cost function $J$ is minimized at every time step. Here $J$ is given by
\begin{align}
	J = \left\Vert{\Delta \mathbf{x}_{t} \shortminus  \Delta \hat{\mathbf{x}}_{t}}\right\Vert_{_2}
\end{align}
where $\left\Vert . \right\Vert_{_2}$ represents the $L^2$ norm.
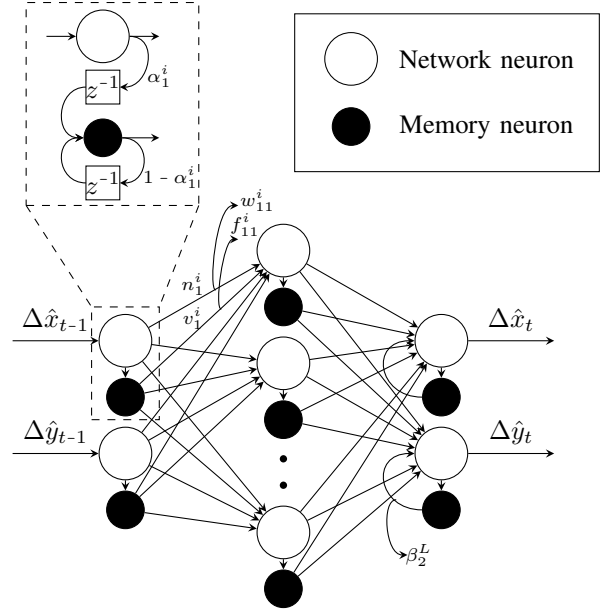
\begin{figure}
\begin{tikzpicture}[x=1.5cm, y=1.5cm, >=stealth]

\node [every neuron](nn-demo) at (0.8, 4.2) {};
\draw [<-] (nn-demo) -- ++(-0.5, 0);
\draw [->] (nn-demo) -- ++(0.5, 0);
\draw (0.65, 3.6) rectangle (0.95, 3.9) node[pos=.5]{$z^{\shortminus 1}$};
\node [neuron memory](mnn-demo) at (0.8, 3.3) {};
\draw [->] (mnn-demo) -- ++(0.5, 0);
\draw (0.65, 2.75) rectangle (0.95, 3.05) node[pos=.5]{$z^{\shortminus 1}$};
\draw [->] (nn-demo.east) .. controls +(left:-4mm) and +(right:4mm) .. (0.95, 3.75) node at (1.3, 3.85) {\scriptsize $\alpha_1^i$};
\draw [->] (0.65, 3.75) .. controls +(left:4mm) and +(right:-4mm) .. (mnn-demo.west);
\draw [->] (mnn-demo.east) .. controls +(left:-4mm) and +(right:4mm) .. (0.95, 2.9) node at (1.4, 2.95) {\scriptsize $1\shortminus\alpha_1^i$};
\draw [->] (0.65, 2.9) .. controls +(left:4mm) and +(right:-4mm) .. (mnn-demo.west);

\draw [dashed] (0.12, 2.7) -- (1.65, 2.7) -- (1.65, 4.5) -- (0.12, 4.5) -- cycle;
\draw [dashed] (0.12, 2.7) -- (0.7, 1.8);
\draw [dashed] (1.65, 2.7) -- (1.3,  1.8);

\node [every neuron](nn-legend) at (3.0, 4.0) {};
\node [neuron memory](mnn-legend) at (3.0, 3.4) {};
\node at (4.2, 4.0) {Network neuron}; 
\node at (4.2, 3.4) {Memory neuron};

\draw (2.5, 4.4) -- (5.2, 4.4) -- (5.2, 3.0) -- (2.5, 3.0) -- cycle;
\foreach \i in {1, 2}
	\node [every neuron/.try](input-\i) at (1.0,2.5-\i) {};

\foreach \i in {1, 2}
	\node [neuron memory/.try](input-memory-\i) at (1.0, 2.0-\i) {};

\foreach \l [count=\i] in {x, y}
	\draw [<-] (input-\i) -- ++(-1, 0)
		node [above, midway] {$\Delta\hat{\l}_{t\shortminus 1}$};
		
\foreach \i in {1, 2}
	\draw[->] (input-\i) -- (input-memory-\i);

\foreach \m [count=\i] in {1, memory, 3, memory, missing, 6, memory}
	\node [every neuron/.try, neuron \m/.try](hidden-\m-\i) at (2.4, 2.8-\i/2) {};

\foreach \i\j in {1/2, 3/4, 6/7}
	\draw[->] (hidden-\i-\i) -- (hidden-memory-\j);

%\foreach \i in {1, 2, 3, 4, 5, 6}
%	\node [neuron memory/.try](hidden-memory-\i) at (2, 5.5-\i*2) {};

\foreach \i in {1, 2}
	\node [every neuron/.try](output-\i) at (3.8,2.5-\i) {};

\foreach \i in {1, 2}
	\node [neuron memory/.try](output-memory-\i) at (3.8, 2.0-\i) {};

\foreach \i in {1, 2}
	\draw[->] (output-\i) -- (output-memory-\i);
	
\foreach \i in {1, 2}
	\foreach \j in {1, 3, 6}
		\draw [->] (input-\i) -- (hidden-\j-\j);

\foreach \i in {1, 2}
	\foreach \j in {1, 3, 6}
		\draw [->] (input-memory-\i) -- (hidden-\j-\j);

\foreach \i in {1, 3, 6}
	\foreach \j in {1, 2}
		\draw [->] (hidden-\i-\i) -- (output-\j);

\foreach \i in {2, 4, 7}
	\foreach \j in {1, 2}
		\draw [->] (hidden-memory-\i) -- (output-\j);

\foreach \i in {1, 2}
	\draw [->] (output-memory-\i.west) .. controls +(left:6mm) and +(right:-6mm) .. (output-\i.west);
	
\foreach \l [count=\i] in {x, y}
	\draw [->] (output-\i) -- ++(1, 0)
		node [above, midway] {$\Delta\hat{\l}_{t}$};

\node at (1.6, 2.0) {\scriptsize $n_1^i$};
\node at (1.6, 1.7) {\scriptsize $v_1^i$};
\draw [->] (1.8, 1.95) .. controls +(left:1mm) and +(right:-3mm) .. (2.0, 2.7);
\draw [->] (1.85, 1.78) .. controls +(left:1mm) and +(right:-2mm) .. (2.0, 2.4);

\node at (2.17, 2.75) {\scriptsize $w_{11}^i$};
\node at (2.06, 2.53) {\scriptsize $f_{11}^i$};

\draw [->] (3.4, 0.09) .. controls +(left:2mm) and +(right:-3mm) .. (3.5, -0.4);
\node at (3.6, -0.4) {\scriptsize $\beta_2^L$};

\draw [dashed] (0.7, 0.8) -- (1.3, 0.8) -- (1.3, 1.8) -- (0.7, 1.8) -- cycle;
\end{tikzpicture}
\centering
\caption{The memory neuron network is fully connected with 6 hidden neurons. Every neuron has a memory neuron associated with it. Initially, the network is trained with zero inputs so that the weights stabilize to some equilibrium point, before providing the actual data.} \label{NN}
\end{figure}
\subsection{Network Architecture}\label{NA}
The network architecture is shown in Fig. \ref{NN}. The figure shows some of the network parameters that provides clarity on understanding the functioning of the network. The Memory Neuron Network consists of fully connected \textit{network neurons} (large open circles) and its associated \textit{memory neurons} (small filled circles). There are weights associated with both the connections of network neurons and memory neurons. Both these weights are updated during backpropagation. 

To describe the functioning of the network, let $\Delta\mathbf{x}_{t\shortminus1} = (\Delta\hat{x}_{t-1}, \Delta\hat{y}_{t-1})$ be the inputs to the network. The net output $n_j^{h}(t)$ of the $j^{th}$ network neuron in the hidden layer $h$ can be calculated as:
\begin{align}
    m_j^h(t) = \sum_{k=1}^{2} w_{kj}^i n_k^i(t) + \sum_{k=1}^{2} f_{kj}^iv_k^i(t)\label{eq:1} \\
    n_j^h(t) = g^h\left( m_j^h(t) \right), \ \ \ \ 1 \leq j \leq 6
\end{align}
where,
\begin{itemize}
\item $w_{kj}^i$ is the weight of the connection from $k^{th}$ network neuron in the input layer $i$ to $j^{th}$ network neuron of the hidden layer $h$.
\item $n_k^i(t)$ is the output of the $k^{th}$ network neuron in the input layer $i$. In our case, $n_1^i(t) = \Delta\hat{x}_{t-1}$ and $n_2^i(t) = \Delta\hat{y}_{t-1}$.
\item $f_{kj}^i$ is the weight of the connection from the memory neuron corresponding to the $k^{th}$ network neuron in the input layer $i$ to $j^{th}$ network neuron of the hidden layer $h$.
\item $v_k^i(t)$ is the output of the memory neuron of the $k^{th}$ network neuron in the input layer $i$.
\item  $g^h(.) = tanh(.)$ is the activation function of the network neurons present in the hidden layer. 
\end{itemize}
The output of the memory neuron corresponding to the $j^{th}$ network neuron in the layer $l$ is given by:
\begin{align}
    v_j^l(t) = \alpha_j^ln_j^l(t\shortminus1) + (1\shortminus\alpha_j^l)v_j^l(t\shortminus1), \ l \in \{ i, h, L \}
\end{align}
where $\alpha_j^l$ is the weight of the connection from $j^{th}$ network neuron in the input layer $l$ to its corresponding memory neuron.
The net output $n_j^L(t)$ of the $j^{th}$ network neuron in the last layer $L$ is calculated as:
\begin{align}
    m_j^L(t) = \sum_{k=1}^6 w_{kj}^hn_k^h(t) + \sum_{k=1}^6 f_{kj}^hv_k^h(t) + \beta_j^Lv_j^L(t) \\
    n_j^L(t) = g^L\left( m_j^L(t) \right), \ \ \ \ 1 \leq j \leq 2 \label{eq:2}
\end{align}
where,
\begin{itemize}
    \item $\beta_j^L$ is the weight of the connection from the memory neuron to its corresponding $j^{th}$ network neuron in the last layer $L$.
    \item $v_j^L(t)$ is the output of the memory neuron corresponding to the $j^{th}$ network neuron in the last layer $L$.
    \item $g^L(.)$ is a linear activation function with unit slope for the network neurons in the output layer $L$.
    \item $n_j^L(t)$ is the output of the $j^{th}$ network neuron in the last layer $L$. In our case, $n_1^L(t) = \Delta\hat{x}_{t}$ and $n_2^L(t) = \Delta\hat{y}_{t}$.
\end{itemize}
To ensure the stability of the network dynamics, the following condition is imposed: $0 \leq \alpha_j^l, \beta_j^L \leq 1$. 

The backpropagation algorithm is used to update all the weights of the network corresponding to both the network neurons as well as the memory neurons. The following squared error function is used for backpropagation: 
\begin{align}
    e(t) = \sum_{j=1}^{2} (n_j^L(t) - d_j(t))^2
    \label{error_calc}
\end{align}
where $d_j(t)$ is the desired teaching signal that is derived from the trajectory database. In our case, $d_1(t) = \Delta x_t$ and $d_2(t) = \Delta y_t$.

At the time of updation $t = \tau$, the weights are updated by using the following rule:
\begin{align}
    w_{kj}^l(\tau+1) = w_{kj}^l(\tau) - \eta e_j^{l+1}(\tau)n_i^l(\tau), \ \ l \in \{i, h\} \label{update:1} \\
    f_{kj}^l(\tau+1) = f_{kj}^l(\tau) - \eta e_j^{l+1}(\tau)v_i^l(\tau), \ \ l \in \{i, h\}
\end{align}
where $\eta$ is the learning rate for the weights of the network, and
\begin{align}
    e_j^L(\tau) = \left( n_j^L(\tau) - d_j(\tau)\right), \ \ 1 \leq j \leq 2 \\
    e_j^h(\tau) = \left(g^h(m_j^i(\tau))\right)'\sum_{p=1}^{2}e_p^L(\tau)w_{jp}^h(\tau), \ \ 1 \leq j \leq 6
\end{align}

The various memory coefficients are updated using the following equations:
\begin{align}
    \alpha_j^l(\tau+1) = \alpha_j^l(\tau) - \eta'\frac{\partial e}{\partial v_j^l}(\tau)\frac{\partial v_j^l}{\partial \alpha_j^l}(\tau) \\ 
    \beta^L_j(\tau + 1) = \beta_j^L(\tau) - \eta'e_j^L(\tau)v_j^L(\tau) \ \ \ \ \
\end{align}
where $\eta'$ is the learning rate for updating the memory coefficients, and
\begin{align}
    \frac{\partial e}{\partial v_j^h}(\tau) = \sum_{s=1}^{N_{l+1}}f^h_{js}(\tau)e_s^L(\tau) \\
    \frac{\partial v_j^l}{\partial \alpha_j^l}(\tau) = n_j^l(\tau \shortminus 1) \shortminus v_j^l(\tau \shortminus 1) \label{update:2}
\end{align}
where $N_{l+1}$ is the number of network neurons in the layer next to $l$. The memory coefficients are hard$\shortminus$limited to $\left[ 0, 1 \right]$ if they happen to fall outside the range. For a detailed discussion on the functioning of the network and additional details, please refer to \cite{sastry1994memory}.

A crucial requirement in system identification problems is to determine how many previous inputs and outputs are to be fed back to the model to capture the generic nonlinear input-output mapping of the model. The presence of the memory neurons ensures that this requirements is optimally learnt during the learning process. Note that the output of the network depends on the previous inputs as well as its own outputs due to the presence of memory neurons in the output layer. Thus, the estimated next state of the system $\Delta \hat{\mathbf{x}}_k$ is given by:

\begin{align}
	\Delta \hat{\mathbf{x}}_{t} = \hat{F}(\Delta \hat{\mathbf{x}}_{t\shortminus 1}, \Delta \hat{\mathbf{x}}_{t\shortminus 2}, ...)
	\label{PIPO}
\end{align}
where $\hat{F}(.)$ is the nonlinear transformation represented by the Memory Neuron Network. The predicted samples $\Delta\hat{\mathbf{x}}_t$ depends on the previous inputs due to the presence of memory neurons in the input and hidden layers, and it depends on its own previous outputs due to the presence of memory neurons in the output layer. Thus, the spatio$\shortminus$temporal look$\shortminus$ahead model represented by Fig. \ref{MNN} is known as parallel identification model \cite{narendra1991identification}.
\begin{table*}
\caption{Root Mean Square Error (RMSE) values (in meters) are reported over a prediction horizon of $5s$ for the NGSIM dataset.}
\label{Table_RMSE}
\begin{tabularx}{\textwidth}{@{}l*{10}{C}c@{}}
\toprule
Time & CV & CV-GMM\cite{deo2018would} & GAIL-GRU\cite{kuefler2017imitating} & LSTM & MATF\cite{zhao2019multi} & CS-LSTM\cite{deo2018convolutional} & S-LSTM\cite{alahi2016social} & UST\cite{he2020ust} & UST-180\cite{he2020ust} & MNN \\
\midrule
$1s$ & $0.73$ & $0.66$ & $0.69$ & $0.68$ & $0.67$ & $0.61$ & $0.65$ & $0.58$ & $0.56$ & $\mathbf{0.36}$ \\
$2s$ & $1.78$ & $1.56$ & $1.56$ & $1.65$ & $1.51$ & $1.27$ & $1.31$ & $1.20$ & $1.15$ & $\mathbf{0.85}$ \\
$3s$ & $3.13$ & $2.75$ & $2.75$ & $2.91$ & $2.51$ & $2.09$ & $2.16$ & $1.96$ & $1.82$ & $\mathbf{1.38}$ \\
$4s$ & $4.78$ & $4.24$ & $4.24$ & $4.46$ & $3.71$ & $3.10$ & $3.25$ & $2.92$ & $2.58$ & $\mathbf{1.92}$ \\
$5s$ & $6.68$ & $5.99$ & $5.99$ & $6.27$ & $5.12$ & $4.37$ & $4.55$ & $4.12$ & $3.45$ & $\mathbf{2.74}$ \\
\bottomrule
\end{tabularx}
\end{table*}

\subsection{Training and Implementation Details}\label{TID}
\begin{algorithm}
    \SetKwFunction{isOddNumber}{isOddNumber}
    % \SetKwInput{Input}{Input}
    % \SetKwInput{Output}{Output}
    \SetKwInOut{KwIn}{Input}
    \SetKwInOut{KwOut}{Output}
    \SetKwInOut{KwInit}{Initialize}

    \KwIn{A list $\mathcal{D} = [d_i]$, $i=1, 2, \cdots, n$, where each element is a set of differential trajectory data $d_i = \left\{\Delta\mathbf{x}_t^{(i)}\right\} = \left\{(\Delta x_t^{(i)}, \Delta y_t^{(i)})\right\}_{t=1}^T$ for vehicle $i$, learning rates $\eta, \eta '$, \textit{epochs}\;}
    %\KwOut{Processed list.}
	\KwOut{Trained memory neuron model for trajectory prediction\;}
    \KwInit{Initialize the weights of the network arbitrarily, except the memory coefficients which are initialized to zero.\;}
    %$newList = [\ ]$

    \ForEach{$d_i \in \mathcal{D}$}{
        \For{$e \leftarrow 0$ \KwTo \textit{epochs}}{
        	\ForEach{$\Delta\mathbf{x}_t \in d_i$}{
        		Compute output of the network using feedforward equations \eqref{eq:1} - \eqref{eq:2};\\
        		Compute error for backpropagation using equation \eqref{error_calc}\\
        		Update all the weights and the memory coefficients using equations \eqref{update:1} - \eqref{update:2};
        	}
        }
    }
    \caption{Training pseudocode}
    \label{MNN_algorithm}
\end{algorithm}

The Trajectory database consists of differences between consecutive trajectory samples, as given by equation \eqref{state_diff}. During the learning process, at every time step $t$ the network receives the previous state information $\Delta \mathbf{x}_{t\shortminus 1}$, and predicts the estimated next state $\Delta \hat{\mathbf{x}}_t$. The actual state of the system $\Delta \mathbf{x}_t$ is then used as a \textit{teaching signal}, to backpropagate the squared error $\left\Vert{\Delta \mathbf{x}_t \shortminus  \Delta \hat{\mathbf{x}}_t}\right\Vert_{_2}^2$ and update both the weights associated with the network neurons and the memory neurons. The network consists of six neurons in the hidden layer, with \textit{tanh(.)} as it's activation function, and \textit{linear} activation function in the output layer. The range of the activation function is adjusted according to the range of the state values of the system, to avoid clipping during the prediction phase. It's slope is also adjusted to provide a linear relationship with unit slope, about the origin.

The entire trajectory data is taken for a vehicle, and the difference between consecutive trajectory samples is calculated and stored in the trajectory database for every vehicle. They will be referred as \textit{differential trajectory samples}. Each sample is then presented to the network sequentially and is trained using backpropagation. One epoch is said to be completed when the last sample in the set of differential trajectories samples is presented and learnt. This procedure is repeated for 1,00,000 epochs, for multiple vehicle trajectories. The learning rates for both type of weights is chosen to be $4\times 10^{-6}$. Algorithm \ref{MNN_algorithm} summarizes the training procedure. The entire model is implemented in \texttt{Python} using \texttt{NumPy} library\cite{harris2020array}.

\section{Performance Evaluation}
In this section, the proposed model is evaluated on two datasets, and the performance is compared quantitatively with several state$\shortminus$of$\shortminus$art techniques by employing the RMSE metric.
\subsection{Datasets}
For evaluating the performance of the proposed model, the following datasets are used:
\begin{enumerate}
\item[(a)]\textit{NGSIM US-101\cite{colyar2007us}}: The Next Generation Simulation (NGSIM) US$\shortminus 101$ dataset consists of trajectory data sampled at $10$Hz, over a span of $45$ minutes. The trajectory data is reported in both global as well as local coordinate frames. These trajectories are recorded from a fixed bird's eye view, and consists of varying traffic conditions. A similar experimental setup is followed as in \cite{deo2018convolutional}, where $3s$ of trajectory history is chosen to predict the estimated trajectories over the horizon of next $5s$ during the testing phase.
\item[]
\item[(b)]\textit{Synthetic Dataset:} In order to predict trajectories of rogue vehicles, the trajectories for 20 different rogue vehicles is generated by using the \texttt{CARLA} simulator. The rogue vehicles are made to skip traffic lights randomly and move in a zig$\shortminus$zag fashion within the lane, while traveling at a dangerously high velocity. They can also change lanes abruptly without any indication. The trajectory data is sampled at $20$Hz over a $1\shortminus$minute duration. In order to capture abrupt changes in the trajectories of rogue vehicles, they are sampled at a higher rate of $20$Hz. The same procedure of choosing $3s$ of trajectory history and predicting the estimated trajectories over the horizon of next $5s$ during the testing phase is followed.
\end{enumerate}
\subsection{Evaluation metric}
During the prediction phase, the differential trajectory samples from the trajectory database is provided for a duration of $3s$ to the network and for the next $5s$, the input to the network is it's previous outputs. The predicted values of the network are summed up with the starting actual trajectory values of each vehicle over the duration of $5s$ to generate the predicted actual trajectory of the vehicle. In order to compare the results of the proposed model quantitatively, the root mean squared error (RMSE) metric is used over all future time steps $T_{_H}$ and number of vehicles $N$:
\begin{equation}
\text{RMSE} = \frac{\sum_{n=1}^N\sqrt{\frac{\sum_{t=1}^{T_{_H}} \left\Vert \mathbf{x}_t^{(n)} - \hat{\mathbf{x}}_t^{(n)} \right\Vert^2}{T_{_H}}}}{N}
\end{equation}
\begin{figure*}
  \centering
  \includegraphics[width=16cm,height=7cm]{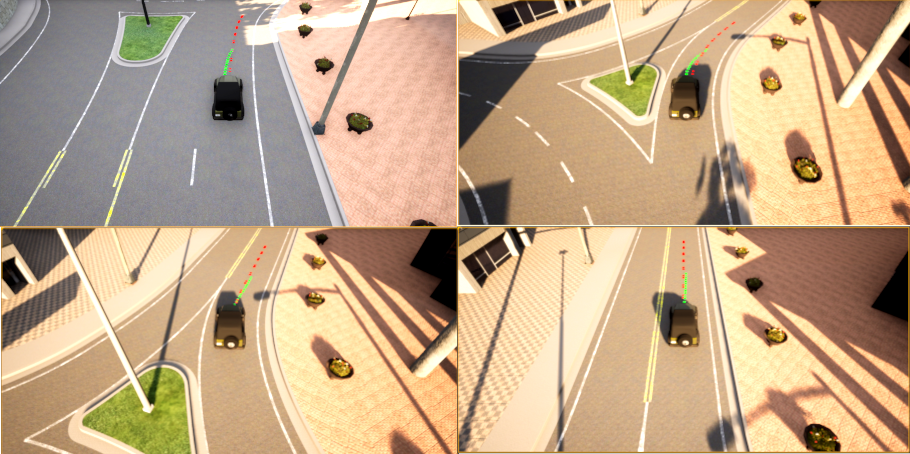}\hfill
  \\[\bigskipamount]\hfill \\
  \centering
  \includegraphics[width=16cm,height=4cm]{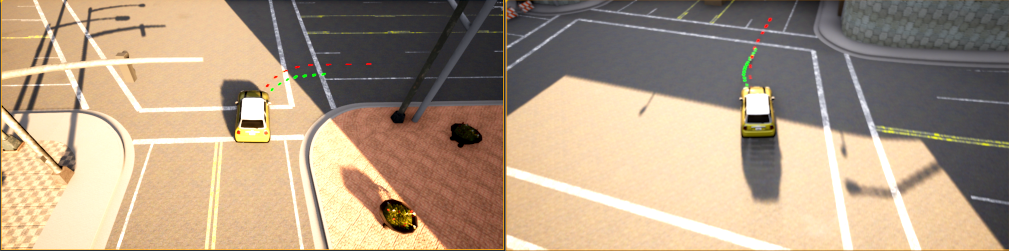}\hfill
  \caption{Simulating trajectory prediction on \texttt{CARLA} for two rogue vehicles. The trained model is deployed on each of the rogue vehicle present in the simulation, so that the other vehicles present in the traffic get a `$5s$' look$\shortminus$ahead of every rogue vehicle. This way, they can plan some protective measures to avoid any collision with them. The predicted trajectories are shown frame-by-frame in green dotted lines for future $5s$, and the actual trajectories given by the planner are shown for $10s$ in red dotted lines. Figure on top shows a car traveling at a roundabout. The bottom figure shows the trajectory prediction at a junction.}
  \label{CARLA_MNN}
\end{figure*}

\subsection{Results}
The performance of the Memory Neuron Network is reported along with several state-of-the-art algorithms tested on the NGSIM US-101 dataset in Table \ref{Table_RMSE}. The table consists of the RMSE for a look$\shortminus$ahead duration of $1s$ to $5s$ for 9 algorithms, which has been reproduced from \cite{he2020ust}. It is evident that the Memory Neuron Network outperforms all the other algorithms. Our results have improved by $35\%$ for $1s$ prediction horizon, and about $20\%$ for $5s$ prediction horizon when compared to \cite{he2020ust}. Further, the rise in the RMSE values from $1s$ horizon to $5s$ horizon is far less for our proposed model, as compared to other algorithms.  From this analysis, it can also be concluded that the proposed model is relatively more stable, than the current existing algorithms. 

This superior performance can be attributed to the fact that the memory neurons not only remember their own past values, but the past values of all the other memory neurons in it's preceding layers as well. This makes the Memory Neuron Network \textit{globally recurrent}, as compared to the LSTM networks which are locally recurrent. 

To test it's robustness, the trained model is deployed in real-time simulation, with 100 cars.\footnote{A detailed video demonstration on the same can be found \href{https://www.youtube.com/watch?v=54DSHSfTy74}{here}.} The simulation is carried out using \texttt{C++ APIs} provided by \texttt{CARLA}'\texttt{s} unreal environment. Only the feedforward part of trained network is implemented in each of the rogue vehicle's trajectory planner.  About $20\%$ of them are rogue vehicles. The simulation consists of mixed vehicles, ranging from small cars to heavy trucks. The future trajectories of all the rogue vehicles are predicted, based on their current location and their $3s$ past track histories. The prediction of the trajectories are shown for two different rogue vehicles as frame-by-frame snapshots in Fig. \ref{CARLA_MNN}.

It can be observed from Fig. \ref{CARLA_MNN} that there is minimal error between the predicted trajectories and the actual future trajectories, when the vehicle is travelling in a near$\shortminus$straight path. The bottom left figure shows the predicted trajectories at the beginning of a left-turn manoeuvre. It is evident that there is a relatively higher error in this scenario, as the model cannot anticipate the radius of curvature of the turning due to the fact that it has no prior knowledge about the map and the dimensions of the roads and junctions present in the map. This shouldn't be concerning, as the predicted trajectory has the same structure of the actual future trajectory, and thus it can be inferred that the vehicle is still going to take a left$\shortminus$turn.

\section{Conclusions and future works}
This paper presents a trajectory prediction model, which uses a novel recurrent neural network as its base model. The trajectory prediction problem is posed as a system identification problem, where the Memory Neuron Network learns the input-output relationship between the past trajectory samples and the future predicted trajectory samples. It is clear that the proposed model outperformed all the state$\shortminus$of$\shortminus$art algorithms currently available, and is also very efficient in the sense that it requires less resources when training, computationally faster due to it's less complicated architecture. The proposed model has a RMSE that is about $20\%$ lesser than the RMSE reported by the current state$\shortminus$of$\shortminus$art algorithms, for a $5s$ look$\shortminus$ahead prediction . The robustness of the proposed model is also verified by deploying it in the \texttt{CARLA} simulator, for each rogue vehicle. While the model performs very well in relatively straighter paths, it fails to predict the trajectories accurately at a junction as it is not aware of the structure of the map. The proposed model will be improved in this regards by adding some features related to the roads and junctions present in the map during the training process, in one of our future works.

\section{Acknowledgments}
The authors would like to thank Dr. Shirin Dora and Dr. Chandan Gautam for their valuable suggestions and comments, and would also like to acknowledge the Wipro$\shortminus$IISc Research Innovation Network (WIRIN) for their financial support.

\bibliographystyle{IEEEtran}
\bibliography{references}
\nocite{*}
\end{document}